\documentclass[10pt, conference, compsocconf]{IEEEtran}
\usepackage{times}
\usepackage{amssymb}
\usepackage{amsmath}
\usepackage{graphicx}
\usepackage{siunitx}
\usepackage{booktabs} 
\usepackage{tikz}
\usepackage{url}

\newcommand\copyrighttext{%
	\footnotesize Copyright 2018 IEEE. Published as a conference paper at the 2018 IEEE Conference 
	on Multimedia Information Processing and Retrieval (MIPR2018) DOI: 
	\url{https://doi.org/10.1109/MIPR.2018.00035}
}
\newcommand\copyrightnotice{%
	\begin{tikzpicture}[remember picture,overlay]
	\node[anchor=south,yshift=10pt] at (current page.south) 
	{\fbox{\parbox{\dimexpr\textwidth-\fboxsep-\fboxrule\relax}{\copyrighttext}}};
	\end{tikzpicture}%
}

\ifCLASSINFOpdf
\else
\fi
\begin{document}
%
\title{Multimodal Image Captioning for Marketing Analysis}


\author{\IEEEauthorblockN{Philipp Harzig, Stephan Brehm, Rainer Lienhart}
\IEEEauthorblockA{University of Augsburg \\
Multimedia Computing and Computer Vision Lab\\
86159 Augsburg, Germany\\
\{philipp.harzig, stephan.brehm, \\ rainer.lienhart\}@informatik.uni-augsburg.de}
\and
\IEEEauthorblockN{Carolin Kaiser, Ren\'{e} Schallner}
\IEEEauthorblockA{GfK Verein\\
90419 Nuremberg, Germany\\
\{carolin.kaiser, rene.schallner \} \\@gfk-verein.org}
}

\maketitle

\begin{abstract}
Automatically captioning images with natural language sentences is an important research topic. State of the art models are able to produce human-like sentences. These models typically describe the depicted scene as a whole and do not target specific objects of interest or emotional relationships between these objects in the image. However, marketing companies require to describe these important attributes of a given scene. In our case, objects of interest are consumer goods, which are usually identifiable by a product logo and are associated with certain brands. From a marketing point of view, it is desirable to also evaluate the emotional context of a trademarked product, i.e., whether it appears in a positive or a negative connotation. We address the problem of finding brands in images and deriving corresponding captions by introducing a modified image captioning network. We also add a third output modality, which simultaneously produces real-valued image ratings. Our network is trained using a classification-aware loss function in order to stimulate the generation of sentences with an emphasis on words identifying the brand of a product. We evaluate our model on a dataset of images depicting interactions between humans and branded products. The introduced network improves mean class accuracy by 24.5 percent. Thanks to adding the third output modality, it also considerably improves the quality of generated captions for images depicting branded products.
\end{abstract}


%
\IEEEpeerreviewmaketitle

\copyrightnotice
\section{Introduction}
In recent years marketing companies have gained lots of experience with text analysis methods by collecting large text corpora from the Internet to identify new trends. One important task is to analyze the interaction with branded products in an emotional context, i.e., marketing companies analyze how people react to and interact with those products in order to track popularity and perception of brands over time. Not only text fragments play an important role in this analysis, but also social media pictures. These images could allow marketing companies to better understand how people interact with certain brands and perceive them emotionally. Consequently, we focus on transferring contents depicted by images into human readable text.

Our contributions are as follows: First, we provide a metric for measuring if an image is correctly classified with respect to objects of interest in the generated caption. Second, we implement a loss function which directly penalizes if the word for the company or brand does not appear in a generated caption. Third, we present a multimodal model which simultaneously generates captions and predicts three different image rating attributes.

\section{Related Work}
Different approaches have appeared over time which try to convert the content of images into text fragments. Li et al. \cite{li2011composing} use image features to compose sentences from scratch by using a text corpus comprised of n-grams gathered from the web. State of the art approaches use modern image classification DCNNs like \cite{szegedy2016rethinking} as feature extractors. One such architecture is the Show and Tell model by Vinyals et al. \cite{vinyals2015show,vinyals2016show}, which makes use of recurrent neural networks (RNNs) for sequence modeling in the form of Long Short-Term Memory (LSTM) networks \cite{hochreiter1997long} to convert image features into human-like sentences. Karpathy et al.~\cite{karpathy2015deep} also utilize neural networks for sequence modeling in form of a Bidirectional RNN to generate captions for images.
Different approaches and their performances are reported on the website for the COCO \cite{lin2014microsoft} captioning challenge. 
The encoder/decoder structure used by current captioning approaches is heavily influenced by machine translation networks that transform a sentence from one language into another. E.g., Google \cite{wu2016google} uses LSTM-driven neural networks for their translation system. Facebook \cite{gehring2017convolutional} goes in another direction and eliminates the LSTM layers by replacing them through ordinary convolutional layers resulting in better performance both in speed and accuracy.

\section{Multimodal Captioning and Rating Model}

Our model aims to produce captions relevant for marketing and to predict image ratings of three different categories from a single image. We therefore enhanced a popular image captioning model with two extensions to allow the combination of the three modalities of image, captions and image ratings in one single neural network architecture.

\subsection{Show and Tell model}

We use the Show and Tell model \cite{vinyals2015show,vinyals2016show} as a basis for our work. This model implements an encoder/decoder network architecture which is inspired by machine translation neural networks and is trained to maximize the probability $p(S|I)$ of generating the target sequence of words $S=(S_1, S_2, \dots )$ for an input image $I$ in an end-to-end fashion. $S$ consists of a variable number of words and the model is optimized for each (image, caption) pair. This model uses an LSTM network as the decoder RNN network and the loss ($L(I, S)$) that defines the error signal is given by the sum over the log likelihoods of the correct word at each position. During training, this loss is minimized w.r.t. (1) the parameters in the LSTM network, (2) the top layer of the CNN, which reduces the feature dimensions to a fixed length vector, and (3) the word embedding matrix ($\textbf{W}_e$), which reduces the feature dimensionality of one-hot vectors representing words in the vocabulary.


\subsection{Dataset}
\label{sec:dataset}

We use a real-world dataset, which was created by a German market research nonprofit association (see Acknowledgment). The dataset consists of images that depict scenarios of persons interacting with branded products, e.g., a man holds a can with a Coca-Cola logo on it. Altogether, this dataset covers 26 different logo classes and consists of 10,529 images. Because of the small number of images in our dataset, we split it into a training and test dataset with a ratio of 9 to 1. We also use 8-fold cross-validation to select the best performing model due to the small number of training images. Two problems arise with our dataset. First, images are distributed unequally among the different classes. Second, we have extremely few training images compared to other datasets like MSCOCO~\cite{lin2014microsoft}. 
Our dataset covers three different modalities. There are 10,529 images contained in our dataset with a total number of 52,645 captions associated with the images. Five annotators created a caption for each image. For a subset of all images (2,718) five annotators created image ratings with five possible values ($0-4$). There are three kinds of image ratings, the first ($r_1$) describing whether the person interacts with the branded product in a positive ($0$) or negative ($4$) way, the second ($r_2$) describing if the person in the image is involved ($0$) with the branded product or uninvolved ($4$) and the third ($r_3$) describing if there is an emotional ($0$) or a functional ($4$) interaction with the branded product. 

\subsection{Classification aware loss for LSTM networks}
\label{sec:cls_aware_loss}

\begin{figure}
	\centering
	\includegraphics[width=0.6\linewidth]{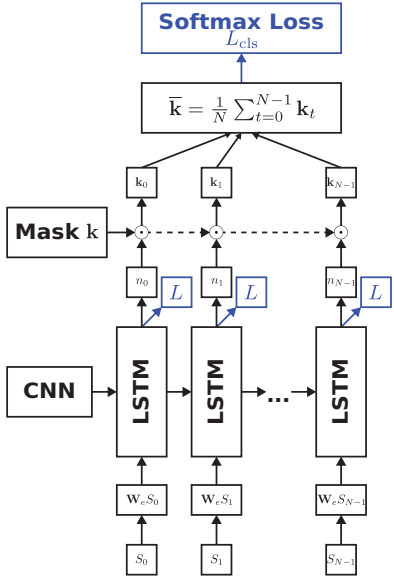}
	\caption{The classification aware loss function on top of the LSTM network.}
	\label{fig:cls_aware}	
	\vspace*{-\baselineskip}
\end{figure}

As we want our model to describe the brand of an object correctly, we need to tell our model, whether or not a generated sentence contains the correct brand. Therefore, we introduce a special loss that punishes if the correct brand name is not contained within a generated sentence. We visualize the classification-aware loss in figure~\ref{fig:cls_aware} with the Show and Tell model on the bottom. We define a classword to be a word in the vocabulary $V$ that clearly identifies a logo class (e.g., the word 'cocacola' is a classword.). We construct a one-hot mask vector $\mathbf{k}$ of size $|V|$, which has a one at every index of a classword. 



$\mathbf{k}_t = \mathbf{n}_t \odot \mathbf{k}$ then provides the scores per word in the vocabulary by only focusing on the classwords, where $\mathbf{n}_t$ is the output of the LSTM at time step $t$. Since a classword can occur at any time step $t$ within the sentence, we average over all words in the sentence ($\overline{\mathbf{k}}$). Note, that $\overline{\mathbf{k}}$ is a vector of size $|V|$ and $pred=\textrm{Softmax}(\overline{\mathbf{k}})$ is the probability distribution over classwords predicted by the generated sentence. Then, we define a cross-entropy loss that produces an error signal at the word for the correct class $C$ with its corresponding one-hot vector $\mathbf{c}$ as $L_{\textrm{cls}}(I,C) = -[\log pred] \cdot \mathbf{c}$. The total loss of our modified model is now given by $L_{\textrm{total}}=L(I,S)+ L_{\textrm{cls}}(I,C)$.


\subsection{Regression of image ratings}

Our third modality are three integer-valued image ratings in the range of $[0, 4]$. We implement this prediction task with a linear regression model. 
To do so, we append a fully-connected layer with one output neuron ($\mathbf{\rho}_r = \mathbf{W}_r \cdot \mathbf{\Theta}$) at the last layer ($\mathbf{\Theta}$) of the \textit{Inception-v3} DCNN for each of the image ratings, where $\mathbf{W}_r$ is a $1\times 2048$ weight matrix for image rating $r$. For the loss function we use the mean squared error function
\begin{equation}
\label{eq:mse}
L_{r}(I) = (\mathbf{\rho}_r - g_r)^2
\end{equation}
with $g_r$ being the ground-truth for image rating $r$. The total loss of our model now changes to $L_{\textrm{total}}=L(I,S)+ L_{\textrm{cls}}(I,C) + L_{r_1}(I) + L_{r_2}(I) + L_{r_3}(I).$ $L_{r_1}(I), L_{r_2}(I) $ and $ L_{r_3}(I)$ are the image rating losses for ratings $r_1,r_2$ and $r_3$, respectively.


\section{Experiments}

In the following, we evaluate our model and compare it with a baseline model, which is trained with no modifications made to a vanilla Show and Tell model.

\subsection{Sentence classification accuracy}

The images of our dataset always show an interaction between a person and an object belonging to a certain company logo (e.g., a person is holding a Coca-Cola bottle). The sentence classification accuracy (SCA) metric measures the accuracy of correctly classified company logos within the generated captions.

For each class, we have a classword in the vocabulary that identifies the class ($C$). We generate three captions for each image using beam search with a beam size of $b=3$. This heuristic considers the $b$ best sentences up to the time step $t$ as candidate sentences for the generation of sentences of length $t+1$. If the classification of an image is \textit{cocacola}, we consider the set of three generated sentences ($\Phi$) a match if the classword is contained in any of them, i.e., $\textrm{match}(I, \Phi, C) = 1$ if the classword for class $C$ is contained in at least one sentence in $\Phi$ and equals to $0$ otherwise. We then calculate the accuracy for every individual class ($\textrm{accuracy}_{C}$) by 
\begin{equation}
\textrm{accuracy}_{C} = \frac{\sum_{I \in \mathbf{I}_{C}} \textrm{match}(I,\Phi,C)}{|\mathbf{I}_{C}|}.
\end{equation}
$\mathbf{I}$ and $\mathbf{I}_{C}$ are the sets of all test images and the test images belonging to class $C$, respectively. We use the mean accuracy (MA) metric because our dataset is distributed unequally among classes (see section~\ref{sec:dataset}).
\begin{equation}
\textrm{MA} = \frac{\sum_{C \in \mathbf{C}} \textrm{accuracy}_{C}}{|\mathbf{C}|},
\end{equation}
where $\mathbf{C}$ is the set of all class labels. We also measure the overall accuracy (OA) by 
\begin{equation}
\textrm{OA} = \frac{\sum_{C \in \mathbf{C}}\sum_{I \in \mathbf{I}_{C}} \textrm{match}(I,\Phi,C)}{|\mathbf{I}|}.
\end{equation}
\subsection{Training configuration}

\paragraph{Inception-v3 network initialization}
Instead of using an \textit{Inception-v3} network pretrained on the 1000 classes of the ImageNet~\cite{russakovsky2015imagenet} classification task, we use an \textit{Inception-v3} network fine-tuned to classify our 26 logo classes.

\paragraph{Show and Tell model} We train the Show and Tell model with the same parameters as Vinyals et al.~\cite{vinyals2016show}. For our extended model with the classification-aware loss, we just optimize all losses simultaneously according to $L_{\textrm{total}}$. 

\paragraph{Image ratings} For the image ratings regression, we reduce the high learning rate of $\eta=2.0$ by a factor of \SI{2000} to $\eta = 0.001$. Because we have image ratings from five different annotators, we train our regression model with the mean of those five ratings as the ground-truth.

\subsection{Results}
\label{sec:results}

In the following, we examine (a) sentence classification accuracy, (b) commonly used metrics for image captioning and (c) predicted image ratings for various models. The baseline model is the Show and Tell model trained on our dataset (\textit{base}). Furthermore, we use the Show and Tell model with the \textit{Inception-v3} model fine-tuned to our 26~logo classes (\textit{base-ft}). Our third model implements the classification-aware loss discussed in section~\ref{sec:cls_aware_loss} and uses the pretrained \textit{Inception-v3} model (\textit{cls-aware}).
We also conduct experiments, in which we merge our dataset with the MSCOCO dataset. To compensate for the fewer images in our dataset we use each image of our dataset eight times to balance both datasets. The resulting ratio is about $0.65 : 1$ for our dataset vs. the MSCOCO dataset. \textit{fuse-1} is the model where both the LSTM part and the regression part are trained on the merged dataset for \SI{1700000} iterations. \textit{fuse-2} continues training \textit{fuse-1} with the parameters of the \textit{Inception-v3} networks unfreezed and a learning rate of $\eta = 0.0005$ untill iteration \SI{3000000}. We do not optimize by the classification-aware loss $L_{\textrm{cls}}(I,C)$ for both \textit{fuse-1} and \textit{fuse-2}. We initialize \textit{fuse-3} with the parameters from \textit{fuse-2}, i.e., \textit{Inception-v3} parameters, LSTM parameters and image rating parameters. \textit{fuse-3} is now trained on our dataset (see section~\ref{sec:dataset}) with the classification-aware loss function enabled (in contrast to \textit{fuse-1} and \textit{fuse-2}).

\paragraph{Sentence classification accuracy}
In table~\ref{tab:sca} we report SCA scores for our three models. We see that our problem mainly profits from fine-tuning the base network to the specific classification labels of our dataset. At a first sight, the classification-aware loss does only marginally improve the SCA scores (\textit{cls-aware}). When using the fused datasets \textit{fuse-1} and \textit{fuse-2}, we notice a small degradation of the SCA scores, which is due to the fact, that the dataset now not only consists out of branded product images, but includes MSCOCO image-caption pairs. \textit{fuse-3} uses all three modalities and allows changes to the \textit{Inception-v3} base network, which as a consequence now allows to improve the SCA scores. The classification-aware loss is enabled in \textit{fuse-3} and improves the mean accuracy in contrast to \textit{cls-aware} by $1.3\%$. The multimodal architecture improves the mean accuracy by $24.5\%$ in total in comparison with \textit{base}.

\paragraph{Bleu-4, Meteor and Cider scores}
In table~\ref{tab:sca} we also list how our models perform according to the well-known BLEU~\cite{papineni2002bleu}, Meteor~\cite{banerjee2005meteor} and CIDEr~\cite{vedantam2015cider} scores. Note, that the scores for \textit{gt} were computed by comparing each annotator against the other four and then averaging over all scores per annotator. For \textit{base}, \textit{base-ft} and \textit{cls-aware} we compared the generated sentence against the five reference sentences. When using the classification-aware loss, all scores decrease compared to the baseline. In contrast to the SCA scores, the fused models profit heavily from the extended dataset, which helps the model to generate more diverse sentences. When finetuning the \textit{Inception-v3} base network (\textit{fuse-2}), the parallel training of the three modalities improve the Bleu-4, Meteor and Cider scores in comparison to \textit{fuse-1}.

\begin{table}
	\centering
	\caption{Sentence classification accuracy and common machine translation metrics for different models on the test dataset.}
	\label{tab:sca}
	\begin{tabular}{lrrrrr}
		model    & OA      & MA      & Cider & Bleu-4 & Meteor \\
		\midrule
		base      & 0.757 & 0.587 & 1.771 & 0.548   & 0.341  \\
		base-ft   & 0.909 & 0.813 & 1.925 & 0.563   & 0.351  \\
		cls-aware & \textbf{0.910} & 0.819 & 1.464 & 0.444   & 0.286  \\
		\midrule
		fuse-1 & 0.889 & 0.796 & 1.858 & 0.542   & 0.344  \\		
		fuse-2 & 0.880 & 0.794 & 2.012 & 0.585   & 0.364  \\
		fuse-3 & 0.908 & \textbf{0.832} & \textbf{2.107} & \textbf{0.612}   & \textbf{0.371}  \\
		\midrule
		gt & \multicolumn{1}{c}{-} & \multicolumn{1}{c}{-} & 1.943 & 0.520   & 0.382  \\
		\bottomrule
	\end{tabular}
\end{table}

\paragraph{Image ratings}

Since we modeled the image ratings as a linear regression model and possible values are in the range $[0, 4]$, we use the mean deviation from the ground-truth as a performance measure. We calculate the deviation for a single example $\epsilon$ with $\text{deviation}_r(\epsilon)  = \sqrt{(\mathbf{\rho}_r(\epsilon) - g_r(\epsilon))^2}$. Note, that this is the square root of the mean squared error as defined in eq.~\ref{eq:mse}. We calculate the mean deviation ($d_r$) for an image rating $r$ by averaging over all examples in the test set. In table~\ref{tab:ratings} we present the calculated deviations on our test set. All image rating deviations increase, when using the fused dataset and start to decrease below the values from \textit{base, base-ft} and \text{cls-aware} when evaluating the \textit{fuse-2} and \textit{fuse-3} models. Note, that training the three modalities in parallel produces best results for all three experiments. The scores for \textit{gt} were computed by comparing each annotator against the other four and then averaging over all deviations per annotator.
\begin{table}
	\centering
	\caption{Deviations from the image ratings on the test set for different models.}
	\label{tab:ratings}
	\begin{tabular}{lrrr}
		model    & $d_{r_1}$   & $d_{r_2}$   & $d_{r_3}$   \\
		\midrule
		base      & 0.5703 & 0.8141 & 0.8504 \\
		base-ft   & 0.5663 & 0.8656 & 0.8078 \\
		cls-aware & 0.5644 & 0.8613 & 0.8054 \\		
		\midrule
		fuse-1    & 0.6071 & 0.9379 & 0.8576 \\
		fuse-2    & \textbf{0.5218} &  \textbf{0.7827} & 0.6974 \\
		fuse-3    & 0.5236 & 0.7876 &  \textbf{0.6968} \\
		\midrule
		gt & 0.2541 & 0.1894   & 0.3138  \\
		\bottomrule
	\end{tabular}
\end{table}

\section{Conclusion}
We proposed an architecture specifically aimed to analyze images relevant for market research. Our model combines the three modalities of image, caption and image ratings into one unified model. We extended an image captioning model to prioritize on describing the brands of products depicted in a scene by using a special loss function, which penalizes if a brand name is not contained within the caption.  Furthermore, we enabled the model to simultaneously infer ratings from the input images. 
We evaluated six different models and merged our dataset with the famous MSCOCO dataset. As a result, we found that extending our specific dataset improves overall sentence quality but decreases the detection rate of brand names. Our approach of combining all three modalities in one model improves caption quality, brand name detection and image ratings and shows that multiple modalities help to get better performance at all tasks of such a model.
\section*{Acknowledgment}
\label{sec:acks}

This work was funded by GfK Verein.
The authors would like to thank Holger Dietrich and Raimund Wildner for the great collaboration.

\bibliographystyle{IEEEtran}
\bibliography{latex8}
%
%

\end{document}